\documentclass[11pt,a4paper]{article}
\usepackage[utf8]{inputenc}
\usepackage[hyperref]{naaclhlt2018}
\usepackage{times}
\usepackage{latexsym}
\usepackage{amsmath}
\usepackage{graphicx}
\usepackage{multirow}
\usepackage{tikz,tikz-qtree}
\usetikzlibrary{positioning,arrows,shapes}
\usepackage{algorithm}
\usepackage{algorithmic}
\usepackage{proof}
\usepackage{gb4ea}
\usepackage{ebproof}
\usepackage{arydshln}
\usepackage[colorinlistoftodos]{todonotes}

\pgfdeclarelayer{background}
\pgfdeclarelayer{foreground}
\pgfsetlayers{background,main,foreground}

\newcommand{\LF}[1]{\ensuremath{\mathbf{#1}}}

\newcommand{\setof}[1]{\ensuremath{\left \{ #1 \right \}}}
\newcommand{\equal}[2]{#1\!=\!#2}
\newcommand{\tuple}[1]{\ensuremath{\left \langle #1 \right \rangle }}

\newcommand{\ndproof}[6][5.5em]{
\tikzstyle{block}=[
text width=#1,
rectangle split,rectangle split parts=2,fill=gray!5,
draw=black!60!white,thick]
\tikzstyle{smallblock}=[
text width=#1,
rectangle split,rectangle split parts=2,fill=gray!5,
draw=black!60!white,thick]
\tikzstyle{arrow}=[thick,->,level distance=30pt,sibling distance=12pt]
\tikzstyle{arrow}=[thick,->,level distance=30pt,sibling distance=12pt]
\begin{tikzpicture}
\node[block] (1) {
\ensuremath{P: #2}
\nodepart{two}
\ensuremath{G: #3}
};
\node[smallblock] (2) [below=0.5cm of 1]{
\ensuremath{P: #5}
\nodepart{two}
\ensuremath{G: #6}
};
\path[arrow] (1) edge node[right]{\sc #4} (2);
\end{tikzpicture}
}

\newcommand{\ndproofsplit}[8][6em]{
\tikzstyle{block}=[
text width=#1,
rectangle split,rectangle split parts=2,fill=gray!5,
draw=black!60!white,thick]
\tikzstyle{smallblock}=[
text width=3.5em,
rectangle split,rectangle split parts=2,fill=gray!5,
draw=black!60!white,thick]
\tikzstyle{arrow}=[thick,->,level distance=30pt,sibling distance=12pt]
\begin{tikzpicture}
\node[block] (1) {
\ensuremath{P: #2}
\nodepart{two}
\ensuremath{G: #3}
};
\node (empty1) [below=0.5cm of 1]{};
\node (empty2) [below=0.1cm of empty1]{};
\node[smallblock] (2) [left=0.0cm of empty2]{
\ensuremath{P: #5}
\nodepart{two}
\ensuremath{G: #6}
};
\node[smallblock] (3) [right=0.0cm of empty2]{
\ensuremath{P: #7}
\nodepart{two}
\ensuremath{G: #8}
};
\path[arrow] (1) edge node[right]{\sc #4} (empty1);
\end{tikzpicture}
}

\tikzset{
var node/.style={circle,draw,node distance=2.0cm}, 
exvar node/.style={dotted,circle,draw,node distance=2.0cm,thick}, 
pred node/.style={rectangle,rounded corners,draw,node distance=1.7cm}, 
rel node/.style={fill=white,circle,inner sep=0.1mm}
}

\newcommand{\redb}{red!55}

\newcommand{\GraphToForm}[1]{#1^{\bullet}}

\newcommand{\phset}[2]{#1(#2)}

\usepackage{url}

\aclfinalcopy 
\def\aclpaperid{565} 

\newcommand{\isa}{\LF{isa}}

\title{Acquisition of Phrase Correspondences using Natural Deduction Proofs}

\author{Hitomi Yanaka$^1$\\ {\tt hitomiyanaka@g.ecc.u-tokyo.ac.jp}\\
		\And
		\hspace{1.5cm}Koji Mineshima$^2$\\
        \hspace{1.5cm}{\tt mineshima.koji@ocha.ac.jp}\\
		\AND
		Pascual Mart\'{i}nez-G\'{o}mez$^3$ \\ {\tt pascual.mg@aist.go.jp}\\
		\And
		\hspace{1.5cm}Daisuke Bekki$^2$\\
        \hspace{1.5cm}{\tt bekki@is.ocha.ac.jp}\\
		\AND
        $^1${\rm The University of Tokyo} \\
        $^2${\rm Ochanomizu University} \\
        $^3${\rm Artificial Intelligence Research Center, AIST} \\
        {\rm Tokyo, Japan}
}
\date{}

\hypersetup{draft}

\begin{document}
\maketitle
\begin{abstract}
How to identify, extract, and use phrasal knowledge is a crucial problem
for the task of Recognizing Textual Entailment (RTE).
To solve this problem, we propose a method for detecting paraphrases via natural deduction proofs of semantic relations between sentence pairs.
Our solution relies on a graph reformulation of partial variable unifications and an algorithm that induces subgraph
alignments between meaning representations.
Experiments show that our method can automatically detect various paraphrases that are absent from existing paraphrase databases.
In addition, the detection of paraphrases using proof information improves the accuracy of RTE tasks.
\end{abstract}


\section{Introduction}
\label{sec:intro}
Recognizing Textual Entailment (RTE) is a challenging natural language processing task that aims to judge whether one text fragment logically follows from another text fragment~\cite{series/synthesis/2013Dagan}.
Logic-based approaches have been successful in representing the meanings of complex sentences, ultimately having a positive impact on RTE~\cite{bjerva:semeval14, beltagy:semeval14,D16-1242, mineshima2016building, abzianidze:2015:EMNLP, abzianidze:2016:*SEM}.
Although logic-based approaches succeed in capturing the meanings of functional or logical words, it is difficult to capture the meanings of content words or phrases using genuine logical inference alone. This remains a crucial problem in accounting for lexical relations between content words or phrases via logical inference.
To solve this problem, previous logic-based approaches use knowledge databases such as WordNet~\cite{Miller:1995:WLD:219717.219748} to identify lexical relations within a sentence pair.
While this solution has been successful in handling word-level paraphrases, its extension to phrase-level semantic relations is still an unsolved problem.
There are three main difficulties that prevent an effective identification and use of phrasal linguistic knowledge.

The first difficulty is the presence of out-of-context phrase relations in popular databases such as the Paraphrase Database (PPDB)~\cite{Ganitkevitch2013}.
PPDB may suggest paraphrases that do not adhere to the context
of our relevant text segments nor to their semantic structure, which might be problematic.

The second difficulty is finding semantic phrase correspondences
between the relevant text segments. Typical approaches only rely on
surface~\cite{beltagy:starsem13} or syntactic correspondences~\cite{arase-tsujii:2017:EMNLP2017}, often producing inaccurate
alignments that significantly impact our inference capabilities. Instead, a mechanism to compute semantic phrase correspondences
could potentially produce, if available, more coherent phrase pairs and solve the recurring issue of discontinuity.

The third difficulty is the intrinsic lack of coverage
of databases for logical inference despite their large size. 
Whereas there is a relatively small number of possible word-to-word correspondences and thus their semantic relations can be enumerated, the same is not true for all phrase pairs that might be of interest.
One alternative is to use functions of infinite domain (e.g., cosine similarity) between phrase representations~\cite{tian-okazaki-inui:2016:P16-1}, but these techniques are still under development, and we have not seen definitive successful applications when combined with logic systems.

In this study, we tackle these three problems.
The contributions of this paper are summarized as follows:
First, we propose a new method of detecting phrase correspondences through natural deduction proofs of semantic relations for a given sentence pair.
Second, we show that our method automatically extracts various paraphrases that
compensate for a shortage in previous paraphrase databases.
Experiments show that extracted paraphrases using proof information improve the accuracy of RTE tasks.

\section{Related Work}
\label{sec:related}
In this section, we review previous logical inference systems that
are combined with lexical knowledge.
The RTE system developed by \citet{abzianidze:2016:*SEM} uses WordNet as axioms
and adds missing knowledge manually from the training dataset; however, this technique requires
considerable human effort and is not extended to handle phrasal knowledge.

\citet{EACL2017} proposed an RTE system with
an on-the-fly axiom injection mechanism guided by a natural deduction theorem prover.
Pairs of unprovable sub-goals and plausible single premises are identified by means of a variable unification routine and then linguistic relations between their logical predicates are checked using lexical knowledge such as WordNet and VerbOcean~\cite{chklovski-pantel:2004:EMNLP}.
However, this mechanism is limited to capturing word-to-word relations within a sentence pair.

\citet{bjerva:semeval14} proposes an RTE system
where WordNet relations 
are used as axioms for word-to-word knowledge in theorem proving.
For phrasal knowledge,
PPDB is used to rephrase an input sentence pair
instead of translating paraphrases into axioms.
However, this solution ignores logical contexts
that might be necessary when applying phrasal knowledge.
Moreover, it does not apply to discontinuous phrases.

\citet{Beltagy:2016:RMC:3068346.3068353} uses
WordNet and PPDB as lexical knowledge in the RTE system.
To increase their coverage of phrasal knowledge, the system combines a resolution strategy to align clauses and literals in a sentence pair and a statistical classifier to identify their semantic relation.
However, this strategy only considers one possible set of alignments between fragments of a sentence pair, possibly causing inaccuracies when there are repetitions of content words and meta-predicates. 

In our research, we propose an automatic phrase abduction mechanism to inject phrasal knowledge during the proof construction process.
In addition, we consider multiple alignments by backtracking the decisions on variable and predicate unifications, which is a more flexible strategy.
We represent logical formulas using graphs, since this is a general formalism that is easy to visualize and analyze. However, we use \emph{natural deduction} (see Section~\ref{sec:nd}) as a proof system instead of Markov Logic Networks for inference.
Some research has investigated graph operations for semantic parsing~\cite{TACL398,reddy_transforming_2016} and
abstractive summarization~\cite{conf/naacl/0004FTSS15}; we contribute to these ideas by proposing a subgraph
mapping algorithm that is useful for performing natural language inferences.

Considerable research efforts have been focused on the identification and extraction of paraphrases.
One successful technique is associated with bilingual pivoting~\citep{Bannard2005, zhao-EtAl:2008:ACLMain2},
in which alternative phrase translations are used as paraphrases at a certain probability.
However, this technique requires large bilingual parallel corpora; moreover, word alignment errors likely cause noisy paraphrases.
Another strategy is to extract phrase pairs from a monolingual paraphrase corpus using alignments between syntactic trees, guided by a linguistically motivated grammar~\cite{arase-tsujii:2017:EMNLP2017}.
The main difference between these studies 
and ours is that they typically attempt alignment between words or syntactic trees, whereas we perform alignments between meaning representations, which
enables the acquisition of more general paraphrases by distinguishing functional words from content words.
This point is important in distinguishing among different semantic relations (e.g., antonyms and synonyms).
In addition, word and syntactic alignments potentially ignore coreferences,
making it difficult to find relations between many-to-many sentences. Semantic alignments enable this because coreferences must refer to the same variable as the original entity.

\section{Logic-based Approach to RTE}
\subsection{Meaning representation}
\label{sec:mr}
In logic-based approaches to RTE,
a text $T$ and a hypothesis $H$ are mapped onto logical formulas $T'$ and $H'$.
To judge whether $T$ entails $H$,
we check whether $T'\Rightarrow H'$ is a theorem in a logical system.

For meaning representations, we use Neo-Davidsonian event semantics~\cite{Parsons90}.
In this approach, a verb is analyzed as a one-place predicate
over events.
Both the arguments of a verb and modifiers are linked to events
by semantic roles, and the entire sentence is closed by
existential quantification over events.
For example, (\ref{ex:init}) is mapped onto (\ref{ex:init-sr}).
\begin{exe}
\ex A girl is skipping rope on a sidewalk. \label{ex:init}
\ex {\small$\exists x_1 \exists x_2 \exists x_3 \exists y_1\, (\LF{girl}(x_1) \wedge \LF{rope}(x_2) \wedge$\\
$\LF{sidewalk}(x_3) \wedge \LF{skip}(y_1) \wedge 
(\equal{\LF{subj}(y_1)}{x_1}) \wedge  (\equal{\LF{obj}(y_1)}{x_2}) \wedge
\LF{on}(y_1, x_3))$} \label{ex:init-sr}
\end{exe}
We use $x_i$ as a variable for entities and $y_j$ for events.
In this semantics, we represent all content words
(e.g., \textit{girl} and \textit{skip})
as one-place predicates.
Regarding functional words,
we represent a preposition like \textit{on} as
a two-place predicate, e.g., $\LF{on}(y_1, x_3)$.
We also use a small set of semantic roles such as
\LF{subj} and \LF{obj} as a functional term
and use equality ($=$) to connect an event and
its participant, as in $\equal{\LF{subj}(y_1)}{x_1}$.

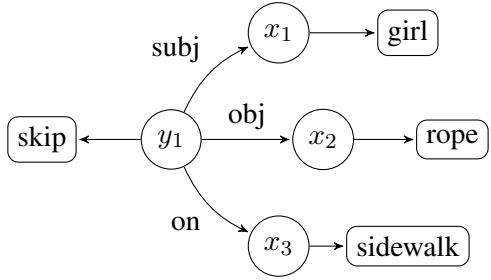
\begin{figure}
\centering
\begin{tikzpicture}
[every edge/.style={->,>=stealth',shorten >=1pt,draw}]
 \node[var node] (y1) {$y_1$};
 \node[pred node] (skip) [left of=y1] {skip};
 \node[var node] (x1) [above right of=y1]  {$x_1$};
 \node[pred node] (girl) [right of=x1] {girl};
 \node[var node] (x2) [right of=y1]  {$x_2$};
 \node[pred node] (rope) [right of=x2] {rope};
 \node[var node] (x3) [below right of=y1]  {$x_3$};
 \node[pred node] (sidewalk) [right of=x3] {sidewalk};
 \path (y1) edge [bend left=20] node [above left] {subj} (x1);
 \path (y1) edge (skip);
 \path (x1) edge (girl);
 \path (y1) edge node [above] {obj} (x2);
 \path (x2) edge (rope);
 \path (y1) edge [bend right=20] node [below left] {on} (x3);
 \path (x3) edge (sidewalk);
\end{tikzpicture}
\caption{A graph for the basic formula (\ref{ex:init-sr}).}
\label{graph-ex}
\end{figure}

To be precise,
the set of \textit{atomic formulas} $\mathcal{A}$ 
in this event semantics
is defined by the rule
\begin{center}
$\mathcal{A} ::= \LF{F}(t) \mid \LF{G}(t,u) \mid t = u$
\end{center}
where $\LF{F}(t)$ is a one-place predicate (for content words), $\LF{G}(t,u)$ is a two-place predicate (for prepositions),
$t$ and $u$ are a term. A term is defined as a constant, a variable, or a functional term of the form $f(t)$
where $f$ is a semantic role and $t$ is a term.

We call a formula constructed by conjunctions and existential quantifiers a \textit{basic formula} in event semantics.
Thus, a set of basic formulas $\varphi$ in event semantics is defined as:
\begin{center}
$\varphi ::= \mathcal{A} \mid \varphi \wedge \varphi \mid \exists t\, \varphi$
\end{center}
The formula in (\ref{ex:init-sr}) is an instance of a basic formula,
which captures the predicate-argument structure of a sentence.

On top of the system of basic formulas, we have
a full language of event semantics
with negation $(\neg)$, disjunction $(\vee)$,
implication $(\to)$, and a universal quantifier ($\forall$).
These operators are used to represent additional logical features.

There is a natural correspondence between basic formulas 
and directed acyclic graphs (DAGs). Figure \ref{graph-ex} shows an example\footnote{
See \citet{jones2016learning} for some variants of
graphical representations of logical formulas. }.
In the graph representation,
constants and variables correspond to vertices;
both two-place predicates for prepositions
(e.g., $\LF{on}(y_1, x_1)$) 
and functional terms for semantic roles (e.g., $\equal{\LF{subj}(y_1)}{x_1}$) are represented as edges.
A one-place predicate $\mathbf{F}(t)$ in a logical formula
can be represented as a functional relation
$\isa(t, \mathbf{F})$, where \isa\ is
an expression relating a term $t$ and a predicate $\LF{F}$
represented as a vertex.
The \isa\ edges are unlabeled for simplicity.

\subsection{Natural deduction and word abduction}
\label{sec:nd}

We use the system of natural deduction~\cite{prawitz1965natural,troelstra2000basic}
to capture phrase correspondences from a sentence pair $(T, H)$,
following the strategies for word axiom injection developed by \citet{EACL2017} and \citet{yanaka2017determining}.
The sentence pair $(T, H)$ is first mapped to a pair of formulas $(T', H')$.
$T'$ is initially set to the premise $P$, and $H'$ is set to the goal $G$ to be proved.

If formulas $P$ and $G$ are basic formulas,
then the proving strategy is to decompose them into a set of atomic formulas using inference rules for conjunctions and existential quantifiers.
The premise $P$ is decomposed into a pool of premises 
$\mathcal{P} = \{\mathbf{p}_i(\theta_i) \mid i\in\setof{1,\dots,m}\}$,
where 
each 
$\mathbf{p}_i(\theta_i)$
is an atomic formula and
$\theta_i$
is a list of terms appearing in 
$\mathbf{p}_i(\theta_i)$.
The goal $G$ is also decomposed into a set of sub-goals 
$\mathcal{G} = \{\mathbf{g}_j(\theta'_j)\mid j\in\setof{1,\dots,n}\}$,
where 
$\theta'_j$
is a list of terms appearing in 
$\mathbf{g}_j(\theta'_j)$.

The proof is performed by searching for a premise 
$\mathbf{p}_i(\theta_i)$
whose predicate matches that of a sub-goal
$\mathbf{g}_j(\theta'_j)$.
If such a premise is found, then 
variables in $\theta'_j$
are unified to 
those in $\theta_i$ 
and the sub-goal 
$\mathbf{g}_j(\theta'_j)$
can be removed from $\mathcal{G}$.
If all the sub-goals can be removed, we prove $T' \to H'$.
In the presence of two or more variables with the same predicate, there might be multiple possible variable unifications.
Modern theorem provers explore these multiple possibilities in search of a configuration that proves a theorem. 

Sub-goals may remain unproved when $T$ logically does not entail $H$
i.e., when there are no premise predicates $\mathbf{p}_i$ that are matched with $\mathbf{g}_j$. In this case,
the system tries word axiom injection,
called \textit{word abduction}.
More specifically,
if there is a premise $\mathbf{p}_i(\theta_i)$
whose predicate has a linguistic relation (according to linguistic knowledge\footnote{As given in a linguistic ontology
or database such as WordNet or VerbOcean.}) with that of a sub-goal $\mathbf{g}_j(\theta'_j)$, then variables in $\theta'_j$ are unified with those in $\theta_i$ and the sub-goal $\mathbf{g}_j(\theta'_j)$ can be removed from $\mathcal{G}$.

\subsection{Graph illustration}
\label{sec:w2w} 
Figure~\ref{graph-outline} shows an example to illustrate
how the system works.
To begin with, the input sentence pair $(T,H)$
is mapped onto a pair of formulas, $(T',H')$.
 $T'$ is initially placed to the premise $P$,
 and $H'$ to the goal $G$.
Note that these are basic formulas,
and they are thus decomposed to the following sets of formulas $\mathcal{P}$ and $\mathcal{G}$, respectively:

\noindent
\scalebox{0.9}{
\begin{tabular}{l}
$\mathcal{P} =\{\LF{lady}(x_1), \LF{meat}(x_2), \LF{cut}(y_1), \LF{up}(y_1),$ \\
\hspace{2.2em} $\LF{precisely}(y_1), \equal{\LF{subj}(y_1)}{x_1}, \equal{\LF{obj}(y_1)}{x_2}\}$ \\
$\mathcal{G} =\{\LF{woman}(x_3), \LF{meat}(x_4), \LF{cut}(y_2),\LF{piece}(x_5),$ \\
\hspace{2.2em} $\LF{into}(y_2, x_5), \equal{\LF{subj}(y_2)}{x_3}, \equal{\LF{obj}(y_2)}{x_4}\}$
\end{tabular}
}

Steps 1 to 3 in Figure~\ref{graph-outline} demonstrate
the variable unification routine and word
axiom injection using graphs.
Note that in step 1, all variables in formulas in $\mathcal{P}$ or
$\mathcal{G}$ are initially different. 

In step 2, we run a theorem proving mechanism that
uses graph terminal vertices as anchors to unify variables between formulas in $\mathcal{P}$ and those in $\mathcal{G}$.
The premise $\mathbf{meat}(x_2)$ in $\mathcal{P}$ matches the predicate $\mathbf{meat}$ of the sub-goal $\mathbf{meat}(x_4)$ in $\mathcal{G}$
and the variable unification $x_4 := x_2$ is applied  
(and similarly for the sub-goal $\mathbf{cut}(y_2)$ in $\mathcal{G}$ with the variable unification $y_2 := y_1$).

In step 3, we use the previous variable unification on $y_1$, the $\LF{subj}$ edge
in $\mathcal{P}$ and $\mathcal{G}$ and the axiom {\small $\forall x. \LF{lady}(x) \to \LF{woman}(x)$}
from external knowledge
to infer that $x_3 := x_1$. 

\section{Phrase Abduction}
\label{sec:method}

There is one critical reason that the word-to-word axiom injection described in Section~\ref{sec:nd} fails to detect phrase-to-phrase correspondences.
That is, the natural deduction mechanism decomposes the goal $G$ into atomic sub-goals that are then proved
\emph{one-by-one} (word-by-word), independently of each other except for the variable unification effect.
This mechanism is particularly problematic when we attempt to prove phrases that resist decomposition,
two-place predicates (e.g., \LF{into}($x,y$)), or failures in variable unification (e.g., due to inaccurate semantics).
Thus, we propose a method to detect phrase-to-phrase correspondence through natural deduction proofs.

\subsection{Phrase pair detection}

\begin{figure*}[ht!]
\small
\centering
\scalebox{0.88}{
\begin{tikzpicture}
[every edge/.style={->,>=stealth',shorten >=1pt,draw},
caption/.style = {right,inner sep=0mm}]

\node[var node] (e1) {$y_1$};
  \node[pred node] (cut) [below left of=e1]  {cut};
  \node[pred node] (up) [below of=e1]  {up};
  \node[pred node] (precisely) [below right of=e1]  {precisely};
  \node[var node] (x1) [above left of=e1] {$x_1$};
  \node[var node] (x2) [above right of=e1] {$x_2$};
  \node[pred node] (lady) [left of=x1] {lady};
  \node[pred node] (meat) [right of=x2] {meat};
  \path (e1) edge [bend right=40] (cut);
  \path (e1) edge (up);
  \path (e1) edge [bend left=40] (precisely);
  \path (e1) edge node [left] {subj} (x1);
  \path (e1) edge node [right] {obj} (x2);
  \path (x1) edge (lady);
  \path (x2) edge (meat);
  \node[exvar node] (e2) [right=7.5cm of e1]{$y_2$}; 
  \node[pred node] (cutr) [below left of=e2]  {cut};
  \node[exvar node] (x3) [above left of=e2] {$x_3$};
  \node[exvar node] (x4) [above right of=e2] {$x_4$};
  \node[exvar node] (x5) [right of=e2] {$x_5$};
  \node[pred node] (woman) [left of=x3] {woman};
  \node[pred node] (meatr) [right of=x4] {meat};
  \node[pred node] (piece) [right of=x5] {piece};
  \path (e2) edge [bend right=40] (cutr);
  \path (e2) edge node [left] {subj} (x3);
  \path (e2) edge node [right] {obj} (x4);
  \path (e2) edge node [below] {into} (x5);
  \path (x3) edge (woman);
  \path (x4) edge (meatr);
  \path (x5) edge (piece);

  \node[text width=7.3cm,align=left] (T) [above=2.6cm of e1] {\normalsize $T$: \textit{A lady is cutting up some meat precisely}};
  \node[text width=8.0cm,align=left] (H) [above=2.6cm of e2] {\normalsize $H$: \textit{Some meat is being cut into pieces by a woman}};
  
  \node[text width=7.3cm,align=center] (f1) [above=1.8cm of e1] {$T': \exists x_1\exists x_2\exists y_1(\LF{lady}(x_1) \land \LF{meat}(x_2) \land  \LF{cut}(y_1) \land \LF{up}(y_1) \land \LF{precisely}(y_1) \land \LF{subj}(y_1,x_1) \land \LF{obj}(y_1, x_2))$};

  \node[text width=8cm,align=center] (f2) [above=1.8cm of e2] {$H': \exists x_3\exists x_4\exists x_5\exists y_2 (\LF{meat}(x_4) \land  \LF{woman}(x_3) \land  \LF{cut}(y_2) \land \LF{piece}(x_5) \land \LF{into}(y_2, x_5) \land \LF{subj}(y_2,x_3) \land \LF{obj}(y_2, x_4))$};
  

  \node[text width=3cm,align=center] (t1) [right=2.2cm of e1] {Step 1: \\ Make graphs from formulas.};
  
  \node[var node] (e12) [below=3.3cm of e1] {$y_1$};
  \node[pred node] (cut2) [below left of=e12]  {cut};
  \node[pred node] (up2) [below of=e12]  {up};
  \node[pred node] (precisely2) [below right of=e12]  {precisely};
  \node[var node] (x12) [above left of=e12] {$x_1$};
  \node[var node] (x22) [above right of=e12] {$x_2$};
  \node[pred node] (lady2) [left of=x12] {lady};
  \node[pred node] (meat2) [right of=x22] {meat};
  \path[very thick,draw=\redb] (e12) edge [bend right=40] (cut2);
  \path (e12) edge (up2);
  \path (e12) edge [bend left=40] (precisely2);
  \path (e12) edge node [left] {subj} (x12);
  \path (e12) edge node [right] {obj} (x22);
  \path (x12) edge (lady2);
  \path[very thick,draw=\redb] (x22) edge (meat2);
  \node[var node] (e1r2) [right=7.5cm of e12,fill=red!20] {$y_1$}; 
  \node[pred node] (cutr2) [below left of=e1r2]  {cut};
  \node[exvar node] (x3r2) [above left of=e1r2] {$x_3$};
  \node[var node] (x2r2) [above right of=e1r2,fill=red!20] {$x_2$}; 
  \node[exvar node] (x5r2) [right of=e1r2] {$x_5$};
  \node[pred node] (womanr2) [left of=x3r2] {woman};
  \node[pred node] (meatr2) [right of=x2r2] {meat};
  \node[pred node] (piecer2) [right of=x5r2] {piece};
  \path[very thick,draw=\redb] (e1r2) edge [bend right=40] (cutr2);
  \path (e1r2) edge node [left] {subj} (x3r2);
  \path (e1r2) edge node [right] {obj} (x2r2);
  \path (e1r2) edge node [below] {into} (x5r2);
  \path (x3r2) edge (womanr2);
  \path[very thick,draw=\redb] (x2r2) edge (meatr2);
  \path (x5r2) edge (piecer2);

  \path[dotted] (cut2) edge [bend right=10] (cutr2);
  \path[dotted] (meat2) edge [bend right=20] (meatr2);
  \path[dotted] (e12) edge [bend right=20] (e1r2); 
\path[dotted] (x22) edge [bend right=20] (x2r2); 

  \node[text width=3cm,align=center] (t2) [right=2.2cm of e12] {Step 2: \\ Anchor terminal vertices and unify variables $x_4 := x_2$ and $y_2 := y_1$.};
  
  \node[var node] (e13) [below=3.3cm of e12] {$y_1$};
  \node[pred node] (cut3) [below left of=e13]  {cut};
  \node[pred node] (up3) [below of=e13]  {up};
  \node[pred node] (precisely3) [below right of=e13]  {precisely};
  \node[var node] (x13) [above left of=e13] {$x_1$};
  \node[var node] (x23) [above right of=e13] {$x_2$};
  \node[pred node] (lady3) [left of=x13] {lady};
  \node[pred node] (meat3) [right of=x23] {meat};
  \path (e13) edge [bend right=40] (cut3);
  \path (e13) edge (up3);
  \path (e13) edge [bend left=40] (precisely3);
  \path[very thick,draw=\redb] (e13) edge node [left] {subj} (x13);
  \path (e13) edge node [right] {obj} (x23);
  \path (x13) edge (lady3);
  \path (x23) edge (meat3);
  \begin{pgfonlayer}{background}
    \draw[rounded corners=2em,line width=3em,blue!15,cap=round]
       (x13.center) -- (lady3.center);
    \end{pgfonlayer}
  \node[var node] (e1r3) [right=7.5cm of e13] {$y_1$};
  \node[pred node] (cut3) [below left of=e1r3]  {cut};
  \node[var node] (x1r3) [above left of=e1r3,fill=red!20] {$x_1$};
  \node[var node] (x2r3) [above right of=e1r3] {$x_2$};
  \node[exvar node] (x53) [right of=e1r3] {$x_5$};
  \node[pred node] (woman3) [left of=x1r3] {woman};
  \node[pred node] (meat3) [right of=x2r3] {meat};
  \node[pred node] (piece3) [right of=x53] {piece};
  \path (e1r3) edge [bend right=40] (cut3);
  \path[very thick,draw=\redb] (e1r3) edge node [left] {subj} (x1r3);
  \path (e1r3) edge node [right] {obj} (x2r3);
  \path (e1r3) edge node [below] {into} (x53);
  \path (x1r3) edge (woman3);
  \path (x2r3) edge (meat3);
  \path (x53) edge (piece3);
  \begin{pgfonlayer}{background}
    \draw[rounded corners=2em,line width=3em,blue!15,cap=round]
       (x1r3.center) -- (woman3.west);
    \end{pgfonlayer}
\path[dotted] (x13) edge [bend left=15] (x1r3);
\path[dotted] (lady3) edge [bend left=15] (woman3);

  \node[text width=3cm,align=center] (t3) [right=2.2cm of e13] {Step 3: \\ Use graph constraints and knowledge (\emph{lady} is a \emph{woman}) to unify $x_3 := x_1$.};

  \node[var node] (e14) [below=3.3cm of e13] {$y_1$};
  \node[pred node] (cut4) [below left of=e14]  {cut};
  \node[pred node] (up4) [below of=e14]  {up};
  \node[pred node] (precisely4) [below right of=e14]  {precisely};
  \node[var node] (x14) [above left of=e14] {$x_1$};
  \node[var node] (x24) [above right of=e14] {$x_2$};
  \node[pred node] (lady4) [left of=x14] {lady};
  \node[pred node] (meat4) [right of=x24] {meat};
  \path[very thick,draw=\redb] (e14) edge [bend right=40] (cut4);
  \path[very thick,draw=\redb] (e14) edge (up4);
  \path[very thick,draw=\redb] (e14) edge [bend left=40] (precisely4);
  \path (e14) edge node [left] {subj} (x14);
  \path (e14) edge node [right] {obj} (x24);
  \path (x14) edge (lady4);
  \path (x24) edge (meat4);
  \begin{pgfonlayer}{background}
    \draw[rounded corners=2em,line width=4.5em,blue!15,cap=round]
       (e14.center) -- (cut4.west) -- (up4.center) -- (precisely4.east) -- (e14.center);
    \end{pgfonlayer}
  \node[var node] (e1r4) [right=7.5cm of e14] {$y_1$};
  \node[pred node] (cutr4) [below left of=e1r4]  {cut};
  \node[var node] (x1r4) [above left of=e1r4] {$x_1$};
  \node[var node] (x2r4) [above right of=e1r4] {$x_2$};
  \node[exvar node] (x5r4) [right of=e1r4] {$x_5$};
  \node[pred node] (womanr4) [left of=x1r4] {woman};
  \node[pred node] (meatr4) [right of=x2r4] {meat};
  \node[pred node] (piecer4) [right of=x5r4] {piece};
  \path[very thick,draw=\redb] (e1r4) edge [bend right=40] (cutr4);
  \path (e1r4) edge node [left] {subj} (x1r4);
  \path (e1r4) edge node [right] {obj} (x2r4);
  \path (e1r4)[very thick,draw=\redb] edge node [below] {into} (x5r4);
  \path (x1r4) edge (womanr4);
  \path (x2r4) edge (meatr4);
  \path (x5r4)[very thick,draw=\redb] edge (piecer4);
  \begin{pgfonlayer}{background}
    \draw[rounded corners=2em,line width=3em,blue!15,cap=round]
       (cutr4.center) -- (e1r4.center) -- (x5r4.center) -- (piecer4.east);
    \end{pgfonlayer}

\node[text width=3cm,align=center] (t4) [right=2.2cm of e14] {Step 4: \\ Induce subgraph alignment with non-unified variable $x_5$.};

\end{tikzpicture}
}
\vspace{-0.4cm}
\caption{A graph representation of a theorem proving routine on basic formulas and variable unification. Dotted circles represent non-unified variables at each step, whereas edges without labels are attributes. The graph of the left side is the set of premises $\mathcal{P}$ and the graph of the right side is the set of sub-goals $\mathcal{G}$.
Colored subgraphs represent a word or a phrase
to which our axiom injection mechanism applies.
}
\label{graph-outline}
\end{figure*}
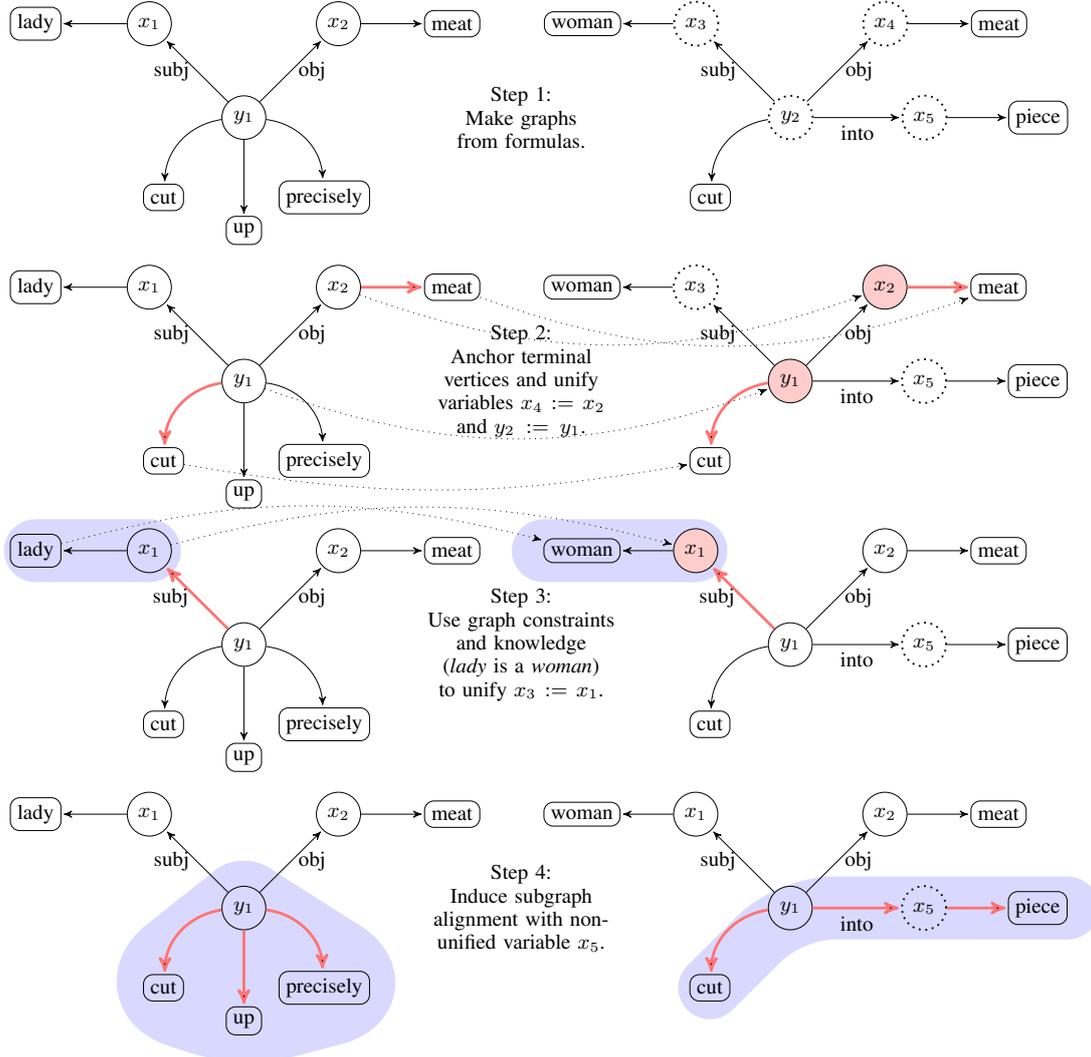
We detect phrase-to-phrase entailing relations between $T'$ and $H'$ by
finding alignments between the subgraphs of their meaning representations when $T' \Rightarrow H'$ or $T' \Rightarrow \neg H'$ hold.
Finding subgraph alignments is a generalization of the subgraph isomorphism
problem, which is NP-complete\footnote{\citet{EMMERTSTREIB2016180} gives a good
overview.}. 
In this paper, we approximate a solution to this problem by
using a combination of a
backtracking variable unification
and a deterministic graph search on the neighborhood
of non-unified variables.

Using our running example in Figure~\ref{graph-outline}, 
step 4 displays our proposed subgraph alignment. The variable $x_5$ in the graph of $\mathcal{G}$
cannot be unified with any variable in the graph of $\mathcal{P}$. This is a very
common case in natural language inferences, as there might be concepts in $H$
that are not directly supported by concepts in $T$. In this research, we propose
spanning a subgraph starting at non-unified variables (e.g., $x_5$ in $\mathcal{G}$)
whose boundaries are semantic roles (e.g., $\LF{subj}$, $\LF{obj}$).
Its candidate semantics from $\mathcal{P}$ are then the attributes of its corresponding
unified variables from $\mathcal{G}$ (e.g. \emph{cut up precisely} $\rightarrow$ \emph{cut into pieces}).

\subsection{Graph alignments}
To formalize this solution we introduce some graph notation.
Let $V = \mathcal{V}^u~\cup~\mathcal{V}^{\bar{u}}~\cup~\mathcal{L}$ be the set of vertices,
where $\mathcal{V}^u$ is the set of unified variables (e.g. $x_1, x_2, y_1$),
$\mathcal{V}^{\bar{u}}$ is the set of non-unified variables (e.g. $x_5$),
and $\mathcal{L}$ is a set of predicates
(e.g., \emph{lady}, \emph{woman}).
Let $E$ be the set of labeled, directed edges $\langle v, l, v'\rangle$ where $v, v' \in V$
and $l$ are labels that may represent a functional relation \isa, 
a preposition or a semantic role.
We denote a set of two-place predicates for prepositions
as \textrm{PREP} and a set of functional terms for
semantic roles as \textrm{ARGS};
e.g., $\textrm{ARGS} = \{\LF{subj}, \LF{obj}\}$.
A graph that represents $\mathcal{P}$ is then a tuple $G_\mathcal{P} = \langle V_\mathcal{P}, E_\mathcal{P} \rangle$,
and similarly, for $\mathcal{G}$, $G_\mathcal{G} = \langle V_\mathcal{G}, E_\mathcal{G} \rangle$.

We can now define a function to span a subgraph in the
neighborhood of non-unified variables $v \in \mathcal{V}_\mathcal{G}^{\bar{u}}$ in the graph of $\mathcal{G}$.
We call a connected set of edges in which no semantic roles appear,
i.e., $\setof{\langle v, l, v'\rangle \mid l \not\in \textrm{ARGS}}$,
a \textit{phrase set}.
Let $\phset{E}{x}$ be the phrase set in $E$
such that each vertex is connected to $x$
with an incoming or outgoing edge, that is,
$
\phset{E}{x} = \setof{(v_i, l, v_k) \in E \mid  (x = v_i \vee x = v_k ) \wedge l \not\in \textrm{ARGS}}.
$
Note that $\phset{E}{x}$ induces a subgraph in a given graph $G$
and the condition $l \notin \textrm{ARGS}$ sets the boundaries of the subgraph by excluding the semantic roles of verb phrases.
Given two phrase sets $E$ and $E'$,
we say $E'$ is reachable from $E$, written $E \sim E'$,
if $E$ and $E'$ share at least one variable vertex.
Let $\sim^*$ be the transitive closure of $\sim$.
Given a set of edges $E_\mathcal{G}$ and a variable $v$,
we define the \textit{extended phrase set},
written $\textrm{Reach}(v)$,
as follows:
\begin{center}
$
\textrm{Reach}(v) = \setof{e \in E \mid \phset{E_{\mathcal{G}}}{v} \sim^* E}
$
\end{center}
that is, the set of edges $e$ that can be reached from $v$ without crossing
an edge with a semantic role label.
This function defines a partition or equivalence class
for non-unified variables $v \in \mathcal{V}_\mathcal{G}^{\bar{u}}$,
and each of these partitions induce a (possibly discontinuous)
phrase in $\mathcal{G}$ that remains unproved.

The corresponding
subgraph in $\mathcal{P}$ to each of these partitions is given by the
vertices and edges connected with a path of length one
to the unified variables that appear in $\textrm{Reach}(v)$.
That is,
%
%
%
\begin{align} \nonumber
\textrm{Corr}(v) = \{ & e \in E_\mathcal{P}(v'), v' \in V_\mathcal{G}^{[v]} \cap V_\mathcal{P} \} \nonumber
\end{align}
where $V_\mathcal{G}^{[v]}$ denotes the vertices in the subgraph of $\mathcal{G}$
induced by the partition $\textrm{Reach}(v)$.

A subgraph alignment between $\mathcal{P}$ and $\mathcal{G}$
is given by the pair of $\langle \textrm{Corr}(v), \textrm{Reach}(v) \rangle$ for all $v \in \mathcal{V}_\mathcal{G}^{\bar{u}}$,
where the phrases can be read from the predicates
in the vertices and edges labeled with prepositions.

We define a mapping $\GraphToForm{(\cdot)}$ from a labeled edge
$\tuple{v, l, v'}$ to an atomic formula as follows.
\begin{center}
$
\GraphToForm{\tuple{v, l, v'}} = 
\left\{
\begin{array}{ll}
    v'(v) & \text{if $l$ is \isa} \\
    l(v, v') & \text{if $l \in \textrm{PREP}$} \\
    l(v) = v' & \text{if $l \in \textrm{ARGS}$}
\end{array}
\right.
$
\end{center}
Let $E$ be a set of labeled edges, and
let $\GraphToForm{E}$ be
$\setof{\GraphToForm{\tuple{v, l, v'}} \mid \tuple{v, l, v'} \in E}$.
The phrase axiom generated for each non-unified variable $v \in \mathcal{V}_\mathcal{G}^{\bar{u}}$
is defined as
\begin{center}
$
\forall \theta_{\textrm{C}}.(\, \bigwedge\GraphToForm{\textrm{Corr}(v)} \to
  \exists \theta_{\textrm{R}}.\,(\bigwedge\GraphToForm{\textrm{Reach}(v)})),
$
\end{center}
where $\theta_{\textrm{C}}$ is a set of free variables appearing in $\GraphToForm{\textrm{Corr}(v)}$ (which includes $v$)
and $\theta_{\textrm{R}}$ is a set of free variables appearing in 
$\GraphToForm{\textrm{Reach}(v)}$ but not in 
$\GraphToForm{\textrm{Corr}(v)}$.

In Figure~\ref{graph-outline},
the only non-unified variable in the sub-goal
in step 4 is $x_5$,
that is, $\mathcal{V}_\mathcal{G}^{\bar{u}} = \setof{x_5}$.
Then, starting from the variable $x_5$,
$\text{Reach}(x_5)$ is
\begin{center}
\small
$\setof{
\tuple{y_1, \LF{into}, x_5},
\tuple{x_5, \isa, \LF{piece}}
}.$
\end{center}
Now $V_\mathcal{G}^{[x_5]} = \setof{y_1, x_5}$, and thus
$\text{Corr}(x_5)$ is
\begin{center}
\small
$\setof{\tuple{y_1, \isa, \LF{cut}},
\tuple{y_1, \isa, \LF{up}},
\tuple{y_1, \isa, \LF{precisely}}
}.$
\end{center}
Finally, the following is the axiom generated from 
$\tuple{\text{Corr}(x_5),\text{Reach}(x_5)}$\footnote{
Note that this axiom is logically equivalent to
\begin{center}
\footnotesize
$\forall y_1 ( \LF{cut}(y_1) 
\land \LF{up}(y_1) \land \LF{precisely}(y_1) 
\to \exists x_5 (\LF{cut}(y_1) 
\land \LF{into}(y_1,x_5) \land \LF{piece}(x_5)))$
\end{center}
indicated in the colored subgraphs in step 4 of
Figure \ref{graph-outline}.}.
\begin{center}
\small
$\forall y_1 ( \LF{cut}(y_1) 
\land \LF{up}(y_1) \land \LF{precisely}(y_1) 
\to \exists x_5 (\LF{into}(y_1,x_5) \land \LF{piece}(x_5)))$.
\end{center}

\subsection{Non-basic formulas}
If formulas $P$ and $G$ are not basic formulas (i.e., 
they contain logical operators other than $\wedge$ and $\exists$), they are decomposed according to inference rules of natural deduction.
There are two types of inference rules: introduction rules decompose a goal formula into smaller sub-goals, and elimination rules decompose a formula in the pool of premises into smaller ones.
Figure \ref{inferencerule} shows introduction rules and elimination rules for decomposing non-basic formulas including negation, disjunction,
implication, and a universal quantifier.
By applying inference rules, a proof of non-basic formulas appearing in sub-goals can be
decomposed to a set of subproofs that only have basic formulas in sub-goals.
If a universal quantifier appears in premises, it is treated in the same way as other premises.

\begin{figure}[t]
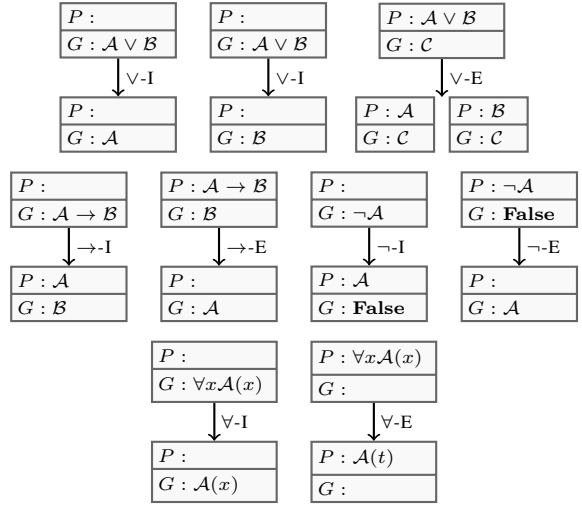

\scriptsize
\centering

\ndproof{}{\mathcal{A \vee B}}
{$\vee$-I}
{}{\mathcal{A}}
\ndproof{}{\mathcal{A \vee B}}
{$\vee$-I}
{}{\mathcal{B}}
\ndproofsplit{\mathcal{A} \vee \mathcal{B}}{\mathcal{C}}
{$\vee$-E}
{\mathcal{A}}{\mathcal{C}}{\mathcal{B}}{\mathcal{C}}

\medskip

\ndproof{}{\mathcal{A} \to \mathcal{B}}
{$\to$-I}
{\mathcal{A}}{\mathcal{B}}
\ndproof{\mathcal{A} \to \mathcal{B}}{\mathcal{B}}
{$\to$-E}
{}{\mathcal{A}}
\ndproof{}{\neg \mathcal{A}}
{$\neg$-I}
{\mathcal{A}}{\LF{False}}
\ndproof{\neg \mathcal{A}}{\LF{False}}
{$\neg$-E}
{}{\mathcal{A}}

\medskip

\ndproof[6.0em]{}{\forall x \mathcal{A}(x)}
{$\forall$-I}
{}{\mathcal{A}(x)}
\ndproof[6.0em]{\forall x \mathcal{A}(x)}{}
{$\forall$-E}
{\mathcal{A}(t)}{}

\caption{Inference rules used for decomposing non-basic formulas.
$P$ is a premise and $G$ is a sub-goal. 
The initial formulas are at the top, with the formulas obtained by applying the inference rules shown below.
}
\label{inferencerule}
\end{figure}
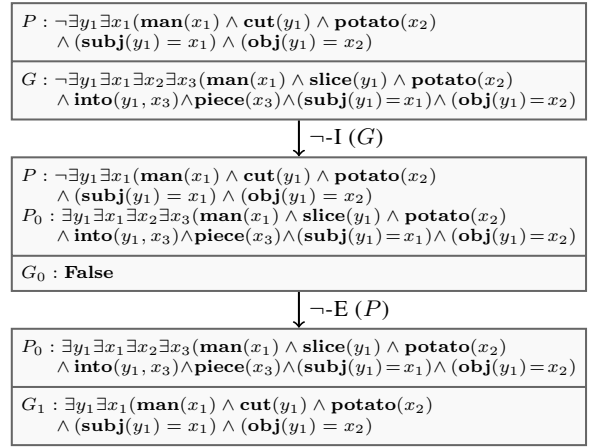
\begin{figure}[t]
\tikzstyle{block}=[
text width=20em,
rectangle split,rectangle split parts=2,fill=gray!5,
draw=black!60!white,thick]
\tikzstyle{smallblock}=[
text width=20em,
rectangle split,rectangle split parts=2,fill=gray!5,
draw=black!60!white,thick]
\tikzstyle{arrow}=[thick,->,level distance=30pt,sibling distance=12pt]
\tikzstyle{arrow}=[thick,->,level distance=30pt,sibling distance=12pt]
\scalebox{0.95}{
\begin{tikzpicture}
\node[block] (0) {
\scriptsize
$P: \neg \exists y_1 \exists x_1(\LF{man}(x_1) \wedge \LF{cut}(y_1) \wedge \LF{potato}(x_2)$\\
$\qquad \wedge \, (\LF{subj}(y_1) = x_1) \wedge (\LF{obj}(y_1) = x_2)$\\
\nodepart{two}
\scriptsize $G: 
\neg \exists y_1 \exists x_1 \exists x_2 \exists x_3 (\LF{man}(x_1) \wedge \LF{slice}(y_1) \wedge \LF{potato}(x_2)$\\
$\qquad \wedge\, \LF{into}(y_1, x_3) \wedge \LF{piece}(x_3) \wedge
(\LF{subj}(y_1) \!=\! x_1) \wedge\, (\LF{obj}(y_1) \!=\! x_2)$\\
};
\node[block] (1) [below=0.5cm of 0]{
\scriptsize
$P: \neg \exists y_1 \exists x_1(\LF{man}(x_1) \wedge \LF{cut}(y_1) \wedge \LF{potato}(x_2)$\\
$\qquad \wedge \, (\LF{subj}(y_1) = x_1) \wedge (\LF{obj}(y_1) = x_2)$\\
$P_0: \exists y_1 \exists x_1 \exists x_2 \exists x_3 (\LF{man}(x_1) \wedge \LF{slice}(y_1) \wedge \LF{potato}(x_2)$\\
$\qquad \wedge\, \LF{into}(y_1, x_3) \wedge \LF{piece}(x_3) \wedge
(\LF{subj}(y_1) \!=\! x_1) \wedge\, (\LF{obj}(y_1) \!=\! x_2)$\\
\nodepart{two}
\scriptsize $G_0: \LF{False}$\\
};
\node[smallblock] (2) [below=0.5cm of 1]{
\scriptsize
$P_0: \exists y_1 \exists x_1 \exists x_2 \exists x_3 (\LF{man}(x_1) \wedge \LF{slice}(y_1) \wedge \LF{potato}(x_2)$\\
$\qquad \wedge\, \LF{into}(y_1, x_3) \wedge \LF{piece}(x_3) \wedge
(\LF{subj}(y_1) \!=\! x_1) \wedge\, (\LF{obj}(y_1) \!=\! x_2)$\\
\nodepart{two}
\scriptsize
$G_1: \exists y_1 \exists x_1(\LF{man}(x_1) \wedge \LF{cut}(y_1) \wedge \LF{potato}(x_2)$\\
$\qquad \wedge \, (\LF{subj}(y_1) = x_1) \wedge (\LF{obj}(y_1) = x_2)$\\
};
\path[arrow] (0) edge node[right]{\footnotesize \sc $\neg$-I ($G$)} (1);
\path[arrow] (1) edge node[right]{\footnotesize \sc $\neg$-E ($P$)} (2);
\end{tikzpicture}
}
\vspace{-0.3cm}
\caption{Proof process for the contradiction.}
\vspace{-0.3cm}
\label{ProveContra}
\end{figure}

For example, consider the following sentence pair with the gold label ``no'' (contradiction):

\noindent $T$: \textit{A man is not cutting a potato}

\noindent $H$: \textit{A man is slicing a potato into pieces}

Figure \ref{ProveContra} shows the proof process of $T' \Rightarrow \neg H'$.
To prove the contradiction,
the formulas $T'$ and $\neg H'$ are set to $P$ and $G$, respectively.
Then, the negation in $G$ is removed by
applying the introduction rule $(\neg\mbox{-}I)$ to $G$.
Here, \LF{False} is the propositional constant denoting the contradiction.
In the second stage of the proof, 
the goal is to prove \LF{False} in $G_0$ from
the two premises $P$ and $P_0$.
By applying $(\neg\mbox{-}E)$ to $P$, we can eliminate the negation from
$P$, resulting in the new goal $G_1$.

As both the premise $P_0$ and the sub-goal $G_1$ are basic formulas,
the procedure described in the previous sections applies to the pair
$(P_0, G_1)$;
these basic formulas are 
decomposed into atomic ones, and then
the word-to-word abduction generates the desired axiom
$\forall y_1 (\LF{cut}(y_1) \!\to\! \LF{slice}(y_1))$.
Finally, the graph alignment applies in the same way as described in Figure~\ref{graph-outline}, which generates the phrase axiom:

\begin{center}
\small
$\forall y_1 ( \LF{cut}(y_1) \to \exists x_5 (\LF{into}(y_1,x_5) \land \LF{piece}(x_5)) )$
\end{center}

\noindent
Using this axiom, one can complete
the proof of the contradiction between $T'$ and $H'$.

\begin{table*}[!t]
\begin{center}
\scalebox{0.90}{
\begin{tabular}{cccc} \hline
   ID & Text & Hypothesis & Entailment\\ \hline \hline
   3941 & \textit{A boy is looking at a calendar} & \textit{There is nobody checking a calendar} & No\\ \hline
   5938 & \textit{Vegetables are being put into a pot by a man} & \textit{Someone is pouring ingredients into a pot}  & Yes \\ \hline
   5930 & \textit{The man is not doing exercises} & \textit{Two men are fighting} & Unknown \\ \hline 
\end{tabular}
}
\vspace{-0.3cm}
\caption{
\label{tab:examples} Examples in the SICK dataset with different entailment labels and similarity scores.}
\end{center}
\end{table*}

\section{Experiments}
\label{sec:experiment}
\subsection{Dataset selection}
We use the SemEval-2014 version of the SICK dataset~\cite{MARELLI14.363} for evaluation.
The SICK dataset is a dataset for semantic textual similarity (STS) as well as for RTE.
It was originally designed for evaluating compositional distributional semantics, so it contains logically challenging problems involving quantifiers, negation, conjunction, and disjunction, as well as inferences with lexical and phrasal knowledge.

The SNLI dataset~\cite{snli:emnlp2015} contains inference problems requiring
phrasal knowledge.
However, it is not concerned with logically challenging expressions; the semantic relationships between a premise and a hypothesis are often limited to synonym/hyponym lexical substitution, replacements of short phrases, or exact word matching.
This is because hypotheses are often parallel to the premise
in structures and vocabularies.
The FraCaS dataset~\cite{cooper1994fracas} also contains logically complex problems.
However, it is confined to purely logical inferences and thus does not contain problems requiring inferences with lexical and phrasal knowledge.
For these reasons, we choose the SICK dataset to evaluate our method of using logical inference to extract phrasal knowledge.

The SICK dataset contains $9927$ sentence pairs with a $5000$/$4927$ training/test split.
These sentence pairs are manually annotated with three types of labels \textit{yes} (entailment), \textit{no} (contradiction), or \textit{unknown} (neutral) (see Table~\ref{tab:examples} for examples).
In RTE tasks, we need to consider a directional relation between words such as hypernym and hyponym to prove entailment and contradiction.
Hence, to extract phrasal knowledge for RTE tasks,
we use the training data whose gold label is \textit{entailment} or \textit{contradiction},
excluding those with the \textit{neutral} label. 

\subsection{Experimental setup}
For the natural deduction proofs, we used ccg2lambda~\cite{martinezgomez-EtAl:2016:P16-4}\footnote{Available at 
https://github.com/mynlp/ccg2lambda.},
a higher-order automatic inference system, which converts CCG derivation trees into semantic representations and conducts natural deduction proofs automatically. 
We parsed the tokenized sentences of the premises and hypotheses using three wide-coverage CCG parsers: C\&C~\cite{clark2007wide}, EasyCCG~\cite{Lewis14a*ccg}, and depccg~\cite{yoshikawa-noji-matsumoto:2017:Long}. CCG derivation trees (parses) were converted into logical semantic representations based on Neo-Davidsonian event semantics (Section~\ref{sec:mr}). The validation of semantic templates used for semantic representations was conducted exclusively on the trial split of the SICK dataset. We used Coq~\cite{opac-b1101046}, an interactive natural deduction theorem prover that we run fully automatically with a number of built-in theorem-proving routines called tactics, which include first-order logic.

\begin{table*}[t]
\begin{center}
\scalebox{0.82}{
\begin{tabular}{cll} \hline
                         Kind&Text &Hypothesis   \\ \hline \hline
   \multirow{2}{*}{noun phrase} & \textit{\textbf{A blond woman} is sitting on the roof of} & \textit{\textbf{A woman with blond hair} is sitting on the roof of} \\ 
   &\textit{a yellow vehicle, and two people are inside}&\textit{a yellow vehicle, and two people are inside}\\ \hline
   \multirow{2}{*}{verb phrase} & \multirow{2}{*}{\textit{The person is \textbf{setting fire to} the cameras}} & \textit{Some cameras are being \textbf{burned} by a person} \\ 
&&\textit{\textbf{with a blow torch}}  \\ \hline
\multirow{2}{*}{verb phrase} &\textit{A man and a woman are \textbf{walking} together} &\textit{A man and a woman are hiking}\\
&\textit{\textbf{through the woods}}&\textit{\textbf{through a wooded area}} \\ \hline
   \multirow{2}{*}{prepositional phrase}   &\textit{A child, who is small, is outdoors climbing}&\textit{A small child is outdoors climbing steps} \\ 
   &\textit{steps outdoors \textbf{in an area full of grass}}&\textit{\textbf{in a grassy area}}\\  \hline
  antonym   &\textit{A woman is \textbf{putting make-up on}}&\textit{The woman is \textbf{removing make-up}}\\  \hline
\end{tabular}
}
\vspace{-0.2cm}
\caption{\label{tab:examples_phrases} Examples of phrase alignments constructed by phrasal axiom injection.}
\vspace{-0.2cm}
\end{center}
\end{table*}

We compare phrase abduction with different experimental conditions.
{\bf No axioms} is our system without axiom injection.
{\bf W2W} is the previous strategy of word abduction~\cite{EACL2017}.
{\bf P2P} is our strategy of phrase abduction;
{\bf W2W+P2P} combines phrase abduction with word abduction.
In addition, we compare our system with three purely logic-based (unsupervised) approaches: {\bf The Meaning Factory}~\cite{bjerva:semeval14}, {\bf LangPro}~\cite{abzianidze:2015:EMNLP}, and {\bf UTexas}~\cite{beltagy:semeval14}.
We also compare our system with machine learning-based approaches: the current state-of-the-art deep learning model {\bf GRU}~\cite{yin-schutze:2017:EACLlong}, a log-linear regression model {\bf SemEval-2014 best}~\cite{lai-hockenmaier:2014:SemEval}, and a hybrid approach combining a logistic regression model and probabilistic logic {\bf PL+eclassif}~\cite{Beltagy:2016:RMC:3068346.3068353}. 

\subsection{Extracted paraphrases}
We extracted 9445 axioms from the SICK training dataset. 
The proving time average to extract phrasal axioms was only 3.0 seconds for a one-sentence pair\footnote{
Ours is a polynomial-time instance of the graph matching problem, where the vertex cover set (maximum number of variables in a phrase) is bounded to a small constant.}.
%
Table~\ref{tab:examples_phrases} shows some examples of paraphrases we extracted from the natural deduction proof in the training set. In particular, the examples of verb phrases show our method has the potential to capture long paraphrases.
Each paraphrase in Table~\ref{tab:examples_phrases} is not contained in WordNet and PPDB.
There are many instances of non-contiguous phrases in the SICK dataset, in particular, verb-particle phrases.
Shown in Table~\ref{tab:examples_phrases}, our semantic alignment can detect non-contiguous phrases through the variable unification process, which is one of the main advantages over other shallow/syntactic methods.
In addition, Table~\ref{tab:examples_phrases} shows our method is not limited to hypernym or hyponym relations, but it is also capable for detecting antonym phrases. 

\begin{table}[!t]
\begin{center}
\scalebox{0.93}{
\begin{tabular}{lccc} \hline
                          & Prec. & Rec.  & Acc.     \\ \hline \hline
   GRU & $-$  & $-$ & $87.1$ \\ \hline                          
   PL+eclassif & $-$  & $-$ & $85.1$ \\ \hline        
   SemEval2014 Best Score & $81.6$  & $81.9$ & $84.6$ \\ \hline \hline 
   The Meaning Factory    & $93.6$  & $60.6$ & $81.6$ \\  \hline  
   LangPro    & $98.0$  & $58.1$ & $81.4$ \\  \hline
   UTexas                 & $-$  & $-$ & $80.4$ \\ \hline \hline
   W2W+P2P             & $84.2$  & $77.3$ & $84.3$ \\  \hline
   W2W           & $97.1$  & $63.6$ & $83.1$ \\  \hline
   P2P             & $85.6$  & $72.1$ & $83.0$ \\  \hline
   No axioms             & $98.9$  & $46.5$ & $76.7$ \\  \hline
\end{tabular}
}
\vspace{-0.3cm}
\caption{\label{tab:sick_rte} RTE results on the SICK dataset.}
\vspace{-0.5cm}
\end{center}
\end{table}

\begin{table*}[th]
\begin{center}
\small
\scalebox{0.72}{
\begin{tabular}{rlccl} \hline
ID&Sentence Pair&Gold&Pred&Axiom \\ \hline \hline
   \multirow{2}{*}{9491} &\textit{A group of four brown dogs are playing in a field of brown grass}& \multirow{2}{*}{Yes} & \multirow{2}{*}{Yes} & $\forall x_1 (\LF{field}(x_1) \wedge \LF{brown}(x_1) \wedge \LF{grass}(x_1)$ \\
                         &\textit{Four dogs are playing in a grassy area}
          &                      &                      & $\rightarrow \LF{grassy}(x_1) \wedge \LF{area}(x_1))$   \\ \hline
   \multirow{2}{*}{2367} &\textit{A person is burning some cameras with a blow torch}& \multirow{2}{*}{Yes} & \multirow{2}{*}{Yes} & $\forall x_1 \forall y_1 (\LF{burn}(y_1) \wedge \LF{with}(y_1, x_1)  \wedge \LF{blow\_torch}(x_1) \wedge \LF{camera}(\LF{obj}(y_1))$ \\
                         &\textit{The person is setting fire to the cameras}
          &                      &                      &
$\rightarrow \LF{set}(y_1) \wedge \LF{fire}(\LF{obj}(y_1)) \wedge \LF{to}(y_1, \LF{obj}(y_1)) \wedge \LF{camera}(\LF{obj}(y_1)))$   \\ \hline
   \multirow{2}{*}{3628} & \textit{A pan is being dropped over the meat} & \multirow{2}{*}{Unk} & \multirow{2}{*}{Yes} & \multirow{2}{*}{$\forall y_1 (\LF{pan}(\LF{obj}(y_1)) \rightarrow \LF{into}(y_1, \LF{obj}(y_1)))$} \\
                         &  \textit{The meat is being dropped into a pan}  &                      &                      &        \\ \hline
   \multirow{2}{*}{96} & \multirow{2}{*}{\begin{tabular}{@{}l@{}}\textit{A man is jumping into an empty pool}
 \\\textit{There is no biker jumping in the air} \end{tabular}} & \multirow{2}{*}{Unk} & \multirow{2}{*}{No} & $\forall y_1 (\LF{jump}(y_1) \rightarrow \exists x_1(\LF{in}(y_1, x_1) \wedge \LF{air}(x_1)))$   \\

                         &                                                                                                                                                  &                      &                      & $\forall y_1 (\LF{man}(y_1) \rightarrow \LF{biker}(y_1))$      \\ \hline
   \multirow{2}{*}{408} &\textit{A group of explorers is walking through the grass}        & \multirow{2}{*}{Yes} & \multirow{2}{*}{Unk} &  \\
                         & \textit{Some people are walking}          &                      &                      &    \\ \hline
\end{tabular}
}
\vspace{-0.2cm}
\caption{\label{tab:examples_neg} Positive and negative examples on RTE from the SICK dataset.}
\vspace{-0.3cm}
\end{center}
\end{table*}

\subsection{RTE evaluation results}
Table~\ref{tab:sick_rte} shows the experimental results.
The results show that the combination of word abduction and phrase abduction improved the accuracy.
When the {\bf W2W+P2P} result is substituted for the {\bf W2W} result, there is a 1.1\% increase in accuracy (from 83.1\% to 84.3\%).
The accuracy of {\bf P2P} is almost equal to that of {\bf W2W}. This is because the recall improves from 63.6\% to 72.1\% while the precision decreases from 97.1\% to 85.6\%. The increase in false positive cases caused this result; some details of false positive cases
are described in the next subsection.
{\bf W2W+P2P} outperformed other purely logic-based systems.
The machine learning-based approaches outperform {\bf W2W+P2P},
but unlike these approaches, parameter estimation is not used in our method.
This suggests that our method has the potential to increase the accuracy by using a classifier.

\subsection{Positive examples and error analysis}
Table~\ref{tab:examples_neg} shows some positive and negative examples on RTE with the SICK dataset.
For ID 9491, the sentence pair requires
the paraphrase from \textit{a field of brown grass} to \textit{a grassy area}, not included in previous lexical knowledges.
Our phrasal axiom injection can correctly generate
this paraphrase from a natural deduction proof, and the system proves the entailment
relation.

ID 2367 is also a positive example of phrasal axiom injection. The phrasal axiom between \textit{set fire to cameras} and \textit{burn cameras with a blow torch} was generated.
This example shows that our semantic alignment succeeds
in acquiring a general paraphrase by separating 
logical expressions such as \textit{some} from content words and also by
accounting for syntactic structures such as the passive-active alternation.

For ID 3628, the axiom shown in the table 
was extracted from the following sentence pair with their \textit{entailment} label:

\noindent $T_1$: \textit{\small A woman is putting meat in a pan}

\noindent $H_1$: \textit{\small Someone is dropping the meat into a pan}

\noindent
But the phrase \textit{drop over} does not entail the phrase \textit{drop into},
and a proof for 
the inference is over-generated in ID 3628.
We extracted all possible phrasal axioms from the training dataset,
so noisy axioms can be extracted
as a consequence of multiple factors such as parsing errors or potential disambiguation in the training dataset. One possible solution for decreasing such noisy axioms would be to use additive composition models~\cite{tian-okazaki-inui:2016:P16-1} and asymmetric learnable scoring functions to calculate the confidence on these asymmetric entailing relations between phrases.

ID 96 is also an example of over-generation of axioms.
The first axiom,
{\small$\forall y_1 (\LF{jump}(y_1) \rightarrow \exists x_1(\LF{in}(y_1, x_1) \wedge \LF{air}(x_1)))$}
was extracted from
the proof of $T_1 \Rightarrow H_1$:

\noindent $T_1$: \textit{\small A child in a red outfit is jumping on a trampoline}

\noindent $H_1$: \textit{\small A little boy in red clothes is jumping in the air}

\noindent
The second axiom {\small $\forall y_1 (\LF{man}(y_1) \rightarrow \LF{biker}(y_1))$}
was extracted from the proof of $T_2 \Rightarrow H_2$:

\noindent $T_2$: \textit{\small A man on a yellow sport bike is doing a wheelie and a friend on a black bike is catching up}

\noindent $H_2$: \textit{\small A biker on a yellow sport bike is doing a wheelie and a friend on a black bike is catching up}

\noindent Although these axioms play a role in the proofs 
of $T_1 \Rightarrow H_1$ and $T_2 \Rightarrow H_2$,
the wrong axiom {\small $\forall y_1(\LF{man}(y_1) \to \LF{biker}(y_1))$} causes the over-generation of a proof for the inference in ID 96.
The correct one would rather be {\small $\forall x_1\forall y_1 (\LF{man}(y_1) \wedge \LF{on}(y_1,x_1) \wedge \LF{bike}(x_1) \to \LF{biker}(y_1))$}. In this case, it is necessary to bundle predicates in a noun-phrase by specifying the types of a variable (entity or event) when making phrase alignments.

For ID 408, the word \textit{explorer} is not contained in the training entailment dataset and hence
the relevant axiom {\small $\forall x_1 (\LF{explorer}(x_1) \rightarrow \LF{people}(x_1))$}
was not generated. While our logic-based method enables detecting semantic phrase correspondences in a sentence pair in an unsupervised way, our next step is to predict unseen paraphrases of this type.

\section{Conclusion}
\label{sec:conc}
In this paper, we proposed a method of detecting phrase correspondences through natural deduction proofs of semantic relations between sentence pairs.
The key idea is to attempt a proof with automatic phrasal axiom injection by the careful management of variable sharing during the proof construction process.
Our method identifies semantic phrase alignments by monitoring the proof of a theorem and detecting unproved sub-goals and logical premises. The method of detecting semantic phrase alignments would be applicable to other semantic parsing formalisms and meaning representation languages such as abstract meaning representations (AMR)~\cite{banarescu-EtAl:2013:LAW7-ID}.
Experiment results showed that our method detected various phrase alignments including non-contiguous phrases and antonym phrases.
This result may contribute to previous phrase alignment approaches.
The extracted phrasal axioms improved the accuracy of RTE tasks.

In future work, we shall enhance this methodology of phrasal axiom injection to predict unseen paraphrases. The pairs of premises and sub-goals that can be detected through the proof process conduct semantic alignments in a sentence pair. With the use of an additive composition model of distributional vectors, we can evaluate the validity of such semantic alignments. A combination of our phrasal axiom injection and additive composition model of distributional vectors has the potential to detect unseen paraphrases in a sentence pair.

\section*{Acknowledgments}
We thank the three anonymous reviewers for their
detailed comments. This work was supported by
JST CREST Grant Number JPMJCR1301 and AIP Challenge Program, Japan.

\bibliography{phrase}
\bibliographystyle{acl_natbib}

\appendix

\end{document}